\documentclass[10pt,twocolumn,letterpaper]{article}

\usepackage[pagenumbers]{cvpr} 

\usepackage{booktabs}
\usepackage[pagebackref,breaklinks,colorlinks]{hyperref}
\usepackage{graphicx}
\usepackage{comment}
\usepackage{amsmath,amssymb} 
\usepackage{booktabs}
\usepackage{multirow}
\usepackage{makecell}
\usepackage{arydshln}
\usepackage[accsupp]{axessibility}

\newsavebox\CBox
\def\textBF#1{\sbox\CBox{#1}\resizebox{\wd\CBox}{\ht\CBox}{\textbf{#1}}}

\DeclareMathOperator*{\argmax}{arg\,max}
\def\etal{\textit{et al}.}
\def\ie{\textit{i.e.}}
\def\eg{\textit{e.g.}}

\usepackage[capitalize]{cleveref}
\crefname{section}{Sec.}{Secs.}
\Crefname{section}{Section}{Sections}
\Crefname{table}{Table}{Tables}
\crefname{table}{Tab.}{Tabs.}

\def\cvprPaperID{8} 
\def\confName{CVPR}
\def\confYear{2023}

\begin{document}

\title{Dynamic Multimodal Fusion}

\author{Zihui Xue \quad Radu Marculescu \\
The University of Texas at Austin\\
}
\maketitle

\begin{abstract}
   Deep multimodal learning has achieved great progress in recent years. However, current fusion approaches are static in nature, \ie, they process and fuse multimodal inputs with identical computation, without accounting for diverse computational demands of different multimodal data. In this work, we propose dynamic multimodal fusion (DynMM), a new approach that adaptively fuses multimodal data and generates data-dependent forward paths during inference. To this end, we propose a gating function to provide modality-level or fusion-level decisions on-the-fly based on multimodal features and a resource-aware loss function that encourages computational efficiency. Results on various multimodal tasks demonstrate the efficiency and wide applicability of our approach. For instance, DynMM can reduce the computation costs by 46.5\% with only a negligible accuracy loss (CMU-MOSEI sentiment analysis) and improve segmentation performance with over 21\% savings in computation (NYU Depth V2 semantic segmentation) when compared with static fusion approaches. We believe our approach opens a new direction towards dynamic multimodal network design, with applications to a wide range of multimodal tasks. \footnote{Our code is available at \url{https://github.com/zihuixue/DynMM}.}
\end{abstract}



\section{Introduction}

Humans perceive the world in a multimodal way, through vision, hearing, touch, taste, etc. Recent years have witnessed great progress of deep learning approaches that leverage data of multiple modalities. Consequently, multimodal fusion has boosted the performance of many classical problems, such as sentiment analysis \cite{sentimentsurvey,sentimentsurvey2,tensorfusion}, action recognition \cite{action,action2}, or semantic segmentation \cite{esanet,cen}.

\begin{figure}[t]
    \centering
    \includegraphics[width=0.8\linewidth]{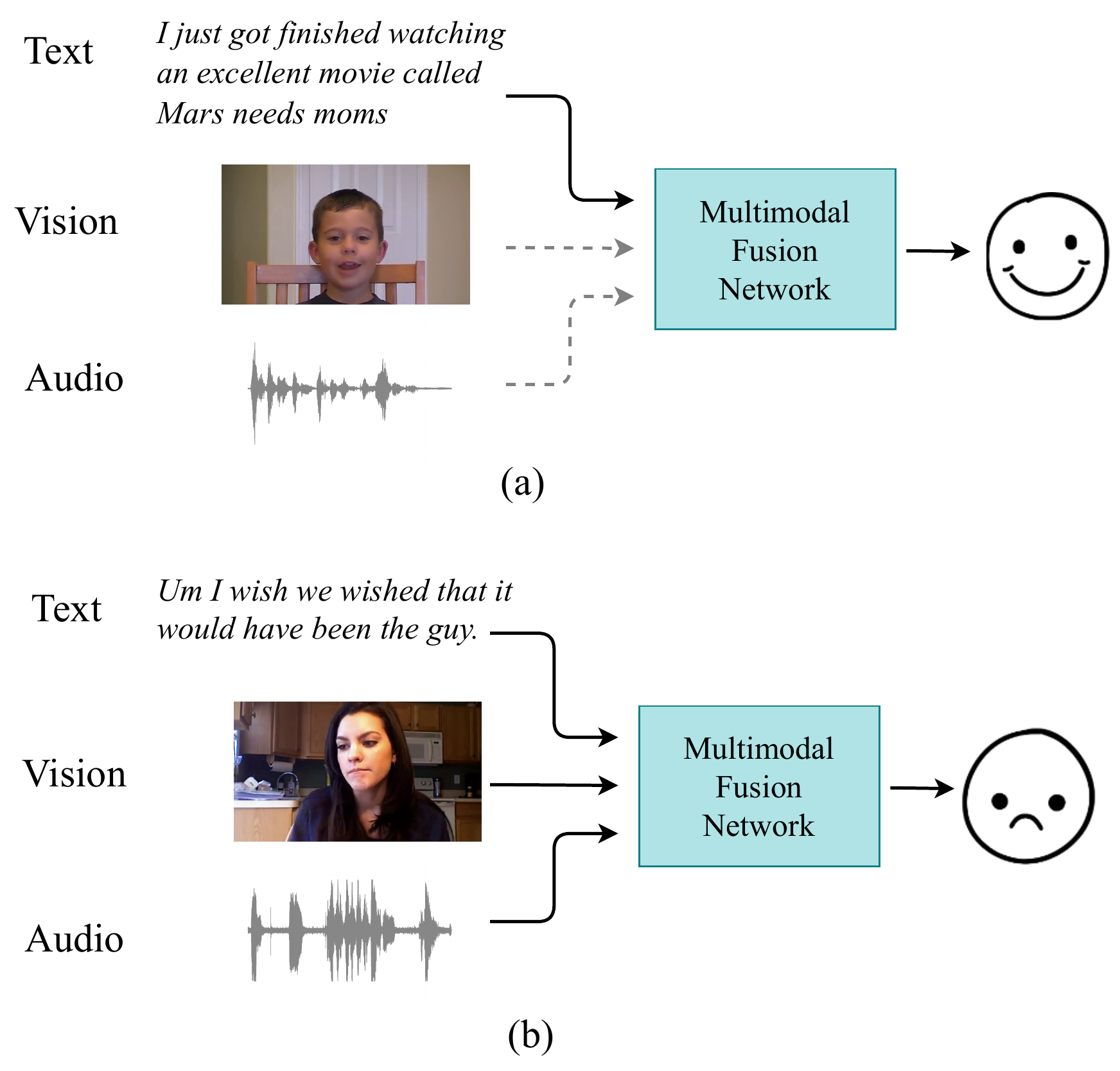}
    \caption{Two examples in CMU-MOSEI \cite{mosei} for emotion recognition. Figure (a) shows an ``easy'' multimodal instance as using textual information is sufficient to predict emotions correctly (this is a positive emotion). Figure (b) shows a ``hard'' example where all three modalities are required to make correct predictions (this is a negative emotion). While static multimodal fusion networks process ``hard'' and ``easy'' inputs identically, we propose \emph{dynamic instance-wise inference} that can achieve computational savings for ``easy'' examples and preserve representation power for ``hard'' instances. For (a), DynMM only activates the text path and skips paths corresponding to the other two modalities, thus leading to computational efficiency.}
    \label{fig.intro}
\end{figure}

Despite these advances, how to best combine information characterized by multiple modalities remains a fundamental challenge in multimodal learning \cite{mmsurvey}. Various research efforts \cite{fusion1,fusion2,fusion3,fusion4,fusion5,fusion6,lrfusion,tensorfusion} have been put into designing new fusion paradigms that can effectively fuse multimodal data. These approaches are generally task- and modality-specific and require manual design. Building on the success of Neural Architecture Search (NAS), a few recent works \cite{mfas,mmnas,mmnas2} have adopted NAS to find effective fusion architectures automatically. 

However, both manually-designed and NAS-based approaches process all the instances in a \emph{single} fusion architecture and lack adaptability to diverse multimodal data. Namely, once the fusion network is trained, it performs static inference on each piece of data, without accounting for the inherent differences in characteristics of different multimodal inputs. Thus, the computational efficiency, as well as the representation power of a well-designed fusion architecture may be limited by its static nature. As a motivating example, consider the two multimodal instances in Figure \ref{fig.intro}. As shown, it is relatively easy to classify emotions for the upper example: the text modality alone provides strong evidence for a positive emotion. On the other hand, it is unlikely to correctly predict emotions for the lower example based solely on the textual information since this sentence is confusing. Audio and visual modalities can provide important cues to a multimodal network to make correct decisions. From this example, we can see that multimodal data enable a model to learn from the rich representations of ``hard'' inputs; it can also bring redundancy in computations for the ``easy'' inputs. 

Inspired by this observation, we propose \emph{dynamic multimodal fusion} (DynMM), a new approach that \emph{adaptively fuses} input data from multiple modalities. Compared with a static multimodal architecture, DynMM enjoys the benefits of reduced computation, improved representation power and robustness. More precisely, dynamic fusion leads to computational savings for ``easy'' inputs that can be correctly predicted using only a subset of modalities or simple fusion operations. For ``hard'' multimodal inputs, DynMM can match the representation power of a static network by relying on all modalities and complex fusion operations for prediction. In addition, real-world multimodal data may be noisy and contradictory \cite{lee2021detect}. In such cases, skipping paths that involve noisy modalities for certain instances in DynMM can reduce noise and boost performance.

Dynamic neural networks~\cite{dynsurvey} have gained increasing attention over the past few years and enjoys a broad range of applications, such as image recognition \cite{wang2020glance,mullapudi2018hydranets,wang2018skipnet,routing}, semantic segmentation \cite{routingseg,dynseg} and machine translation \cite{shazeer2017outrageously}. Motivated by the great success of dynamic inference for unimodal networks, this paper aims at proposing multimodal fusion as a new application domain. To this end, we draw inspiration from the natural redundancy of multimodal data, which provides a different angle from existing work. To be specific, we propose \emph{progressive fusion}, both at \emph{modality level} and at \emph{fusion level}. 
At modality level, we train a gating network to select a subset of input modalities (or all modalities) for predictions based on each input. At fusion level, the gating network provides sample-wise decisions on which fusion operation to adopt and when to stop fusion. On one hand, by allowing exits at the early fusion stages for ``easy'' inputs, DynMM saves the computations of executing the later fusion modules. On the other hand, in terms of ``hard'' multimodal inputs, DynMM can turn all fusion modules on for accurate predictions. 

To verify the efficacy and generalizability of our approach, we conduct experiments on various popular multimodal tasks. DynMM strikes a good balance between computational efficiency and learning performance. For instance, for RGB-D semantic segmentation tasks, DynMM achieves a +0.7\% mIoU improvement with over 21\% reductions in multiply-add operations (MAdds) for the depth encoder when compared against \cite{esanet}. Moreover, we find that DynMM yields better predictions than static fusion networks when the input modality is perturbed by noise; this suggests possible use of DynMM to improve the multimodal robustness. 




\section{Related Work}\label{sec.related}

\subsection{Dynamic Neural Networks}
Dynamic neural networks have demonstrated a great potential in classical computer vision problems, such as image classification \cite{wang2020glance,mullapudi2018hydranets,wang2018skipnet,routing}, object detection \cite{detection1,detection2}, or semantic segmentation \cite{routingseg,dynseg}. While popular deep learning approaches perform inference in a static manner, dynamic networks allow the network structure to adapt to the input characteristics during inference. This flexibility yields many benefits, including high efficiency, representation power and results interpretability \cite{dyn1,dyn2,dyn3}. Dynamic network designs can be categorized into: (a) dynamic depth; (b) dynamic width; (c) dynamic routing \cite{dynsurvey}.

The idea of dynamic depth is to adjust the network depth based on each sample. By providing early exits~\cite{earlyexit1,earlyexit2} in shallow layers, one can save computations by not activating deep layers for ``easy'' samples. For dynamic width, the idea is to adapt the network width in a sample-wise manner. To build a dynamic width network and achieve inference efficiency, previous works have proposed to skip neurons in fully-connected layers \cite{bengio2015conditional}, skip branches in Mixture-of-Experts (MoE) \cite{mullapudi2018hydranets,shazeer2017outrageously}, or skip channels in Convolutional Neural Networks (CNNs) \cite{huang2018condensenet}. To enable more flexibility, recent works \cite{routing,routingseg} build SuperNets with multiple inference paths. Dynamic routing is thus performed inside the SuperNet to generate data-dependent forward paths during inference. Our proposed modality-level DynMM belongs to the category of \emph{dynamic width} approaches; the fusion-level DynMM can be seen as a \emph{dynamic routing} approach.

\subsection{Multimodal Learning}

Multimodal fusion networks have a clear advantage over their unimodal counterparts in various applications, such as sentiment analysis \cite{sentimentsurvey,sentimentsurvey2,tensorfusion}, action recognition \cite{action,action2}, or semantic segmentation \cite{esanet,cen,sagate}. However, how to effectively combine multimodal features to better exploit information remains a big challenge. Existing works either propose hand-crafted fusion designs based on domain knowledge \cite{fusion3,fusion4,fusion5,fusion6,lrfusion,tensorfusion}, or apply NAS to find good architectures automatically \cite{mfas,mmnas,mmnas2}. However, the scope of these works is limited to static networks only.

There have been some early attempts in adopting dynamic neural networks for multimodal applications, such as semantic segmentation~\cite{cen}, video recognition~\cite{gao2020listen,panda2021adamml}, visual-inertial odometry~\cite{yang2022efficient} and medical classification~\cite{han2022multimodal}. Among them, CEN~\cite{cen} dynamically exchanges channels between sub-networks of the RGB and depth modality for performance improvement. Han \etal~\cite{han2022multimodal} proposes to dynamically evaluate feature-level and modality-level informativeness of different samples for more trustworthy medical classification, yet the angle of computational efficiency brought by the dynamic neural networks is overlooked. The work of Gao \etal~\cite{gao2020listen} and AdaMML~\cite{panda2021adamml} are most relevant to our approach as they also adaptively utilize modalities for efficient video recognition. However, their methods are tailored for video data and action recognition. In this work, we aim to make the first step towards a systematic and general formulation of dynamic multimodal fusion that can suit various multimodal tasks.


\section{Method} \label{sec.method}
In this section, we present the key design contributions of our proposed dynamic multimodal fusion network (DynMM). First, we introduce new decision making schemes that enable DynMM to generate data-dependent forward paths during inference. Two levels of granularity are considered, \ie, modality-level (coarse level) and fusion-level (fine level) decision making. Next, we propose new training strategies for DynMM, which consist of (1) a training objective that accounts for resource budgets, and (2) optimization of a non-differentiable gating network.

\subsection{Modality-level Decision}

Assume that input data has $M$ modalities, denoted by $\mathbf x=(x_1,x_2,\cdots,x_M)$. Following the classical Mixture-of-Experts (MoE) \cite{masoudnia2014mixture} framework, we design a set of expert networks as follows. Each expert specializes in a subset of all $M$ modalities. If $M=3$, for example, we can have up to 7 expert networks, denoted by $E_1(x_1)$, $E_2(x_2)$, $E_3(x_3)$, $E_4(x_1, x_2)$, $E_5(x_2, x_3)$, $E_6(x_1, x_2)$, $E_7(x_1, x_2, x_3)$. In real applications, the candidate expert networks can be narrowed down with domain expertise. For instance, depth images can provide useful cues when combined with RGB images, but often perform poorly by themselves in semantic segmentation. In such a case, we do not consider adopting an expert network that only takes depth as input.

Let $B$ represent the number of expert networks that get selected. We propose a \emph{gating network}, denoted by $G(\mathbf x)$, to decide which expert network should be activated. This gating network takes multimodal inputs $\mathbf x$ to form a global view and then produces a $B$-dimensional sparse vector $\mathbf g$ as output. The final output $y$ takes the form of: $y = \sum_{i=1}^B g_i E_i(\mathbf x_i)$, where $\mathbf x_i$ denotes the subset of modalities that the $i$-th expert takes as input.

\begin{figure}[!t]
    \centering
    \includegraphics[width=0.9\linewidth]{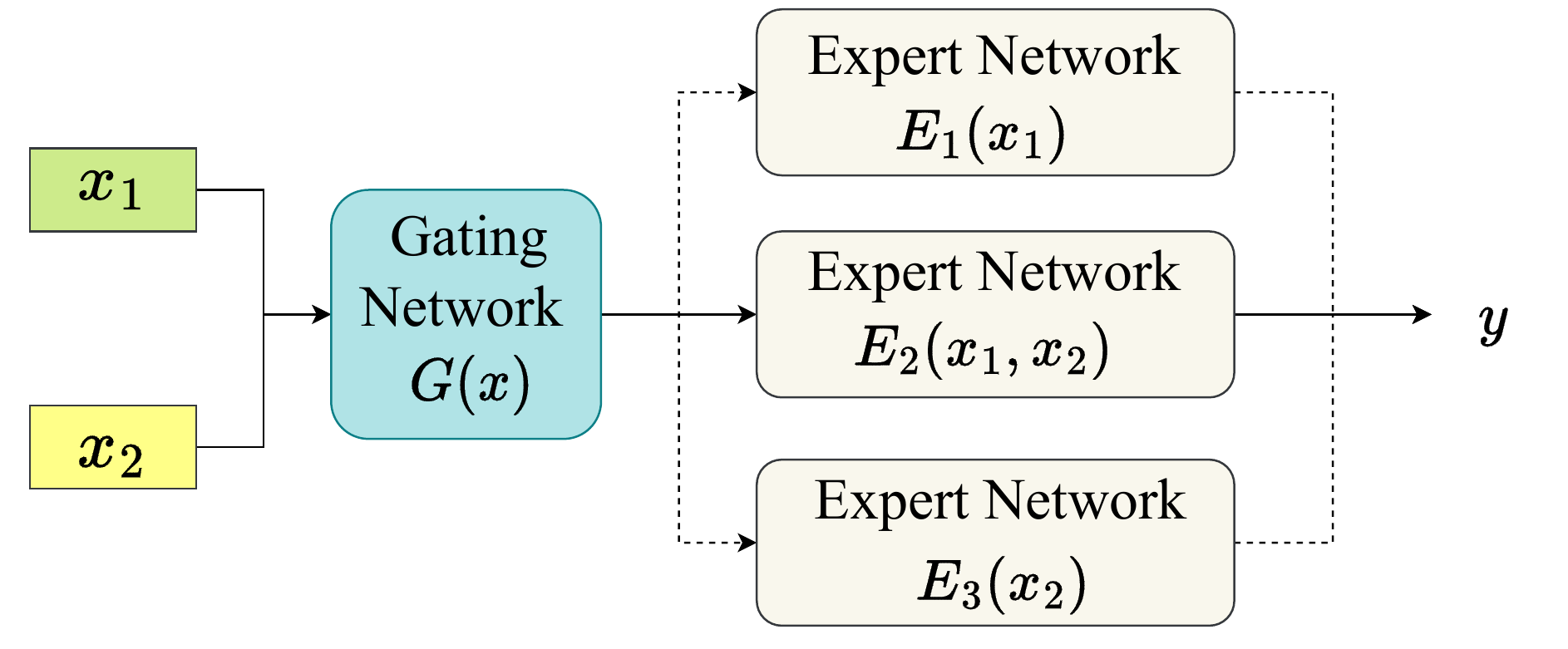}
    \caption{An illustration of modality-level DynMM, where input data has two modalities, denoted by $x_1$ and $x_2$, and the output is denoted by $y$. We design a set of expert networks $\{E_i\}$ that specialize in different subsets of modalities and adopt a gating network $G(\mathbf x)$ to generate data-dependent decisions on which expert network to select.}
    \label{fig.moe}
\end{figure}

Different from conventional MoEs \cite{masoudnia2014mixture} where the output is a weighted summation of expert networks and every branch is executed, in our formulation, the output of the gating network $\mathbf g$ is a one-hot encoding, \ie, only \emph{one} branch is selected for each instance. Therefore, the computations required for other expert networks can be saved. Note that since our expert network already covers a broad range of modality combinations, we only select one branch (as opposed to say selecting top $K$ branches) during each forward pass for maximum computational savings. Figure \ref{fig.moe} provides an illustration of the proposed design with 2 modalities and 3 expert networks (\ie, $M=2$ and $B=3$). 

\begin{figure*}[!t]
    \centering
    \includegraphics[width=1.0\linewidth]{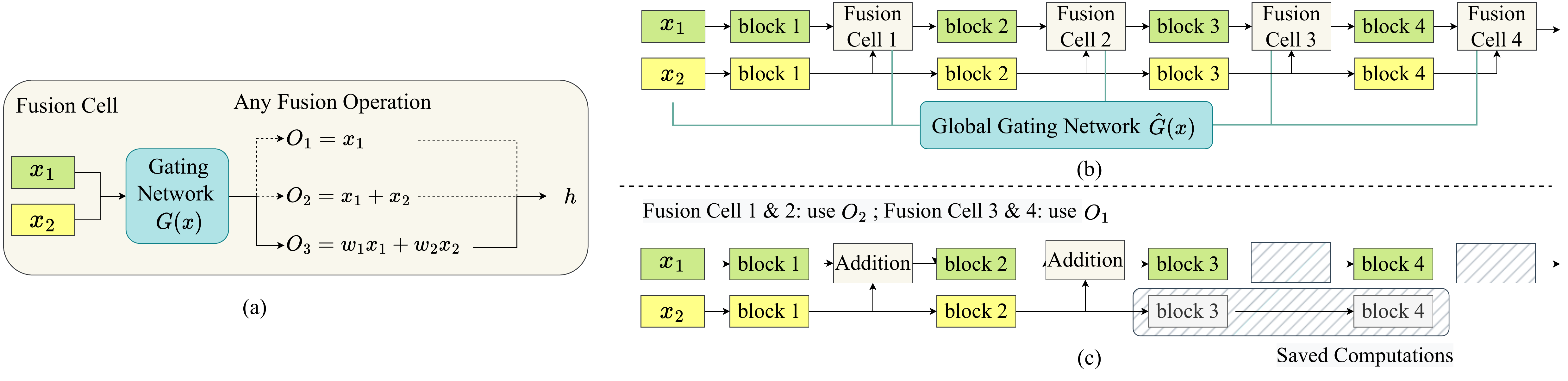}
    \caption{(a) An illustration of fusion-level DynMM, where input data has two modalities, denoted by $x_1$ and $x_2$. We design a fusion cell with a set of candidate operations $\{O_i\}$ and a gating network $G(\mathbf x)$. $h$ represents output of the cell. (b) A dynamic multimodal architecture with stacked fusion cells, where we interlace static feature extraction blocks (colored with green and yellow) with dynamic fusion cells. Gating network $G(\mathbf x)$ in four fusion cells are integrated as one global gating network $\hat G(\mathbf x)$ that outputs decisions for four cells at once. (c) An example architecture when the gating network chooses $O_2$ for the first 2 fusion cells and $O_1$ for the last 2 cells. Consequently, computations of fusion cell 3 \& 4 and feature extraction cell 3 \& 4 for $x_2$ are saved.}
    \label{fig.fusion}
\end{figure*}

The design of the gating network $G(\mathbf x)$ follows two general requirements: (1) it should be computationally cheap to have a small overhead (2) it needs to be sufficiently expressive to make informative decisions on which expert to select. Various gating networks have been proposed previously; they are usually tailored for specific tasks and network architectures~\cite{dynsurvey}. In the experiments, we consider different gating networks (\ie, a multi-layer perceptron (MLP) gate, a transformer gate and a convolutional gate) for three multimodal tasks and provide the detailed description of our gating network architecture in Sec. \ref{sec.exp}.

One remaining problem is the training of gating network $G(\mathbf x)$. Due to the non-differentiability of the discrete decisions given by $G(\mathbf x)$, the network can not be directly trained with back-propagation. Thus, we propose reparameterization techniques and discuss them later in Sec. \ref{sec.train}.


Finally, this gating network $G(\mathbf x)$ is not restricted to taking input-level features; it can also take intermediate features per modality as inputs. Thus, modality-level DynMM can be plugged into any part of a multimodal network and achieve savings in computations after this gating network. 

\subsection{Fusion-level Decision}
While the modality-level decisions directly impact the computational efficiency, completely skipping computations of one modality will likely lead to a downgraded performance for some challenging tasks, \eg, semantic segmentation. Thus, we provide a finer-grain formulation of DynMM with fusion-level decisions next.

We first present the design of a \emph{fusion cell}. Assume input data has $M$ modalities, \ie, $\mathbf x=(x_1,x_2,\cdots,x_M)$. Denote a set of fusion operations as $\{O_i\}$. $O_i$ can be implemented as any function to fuse multimodal features, such as simple identity mapping (\ie, $O_i=x_1$), addition (\ie, $O_i=x_1+x_2+\cdots+x_M$), concatenation (\ie, $O_i=[x_1, x_2,\cdots,x_M]$) and self-attention. Figure \ref{fig.fusion} (a) presents an example design of the fusion cell with two input modalities (\ie, $\mathbf x =(x_1, x_2)$) and three operations (\ie, $O_1=x_1$, $O_2=x_1 + x_2$, $O_3=w_1x_1+w_2x_2$), where $w_1$ and $w_2$ are learnable parameters. Note that here we simplify the operation set for illustration; in practice, we can always adopt more complex fusion operations in each cell to enlarge the representation power. Let $B$ represent the total number of operations. A gating network $G(\mathbf x)$ takes multimodal inputs and produces a $B$-dimensional vector $\mathbf g$ that decides which operation to execute. The output of cell $h$ can be represented as: $h = \sum_{i=1}^B g_i O_i(\mathbf x)$. Following the previous discussion, we adopt hard gates (\ie, $\mathbf g$ is one-hot) for computational efficiency.


Fusion-level DynMM allows decisions at a finer granularity and in a more flexible way by stacking fusion cells to build a dynamic network. We provide an example architecture in Figure \ref{fig.fusion} (b) that we use in our experiments for semantic segmentation ($x_1$ and $x_2$ denote RGB and depth images, respectively). The network consists of four fusion blocks and a global gating network, which allows us to flexibly control the degree of fusion in a sample-wise manner. For instance, we show the resulting architecture in Figure \ref{fig.fusion} (c), when the gating network selects $O_2$ for fusion cell 1 \& 2, and $O_1$ for fusion cell 3 \& 4. This not only skips complex fusion operations that are not selected within the fusion cell, but also saves unnecessary computations in the feature extraction layer. Since we only adopt features from modality 1 after fusion cell 2, there is no need to further process features from modality 2. Thus, we can skip computations in the feature extraction layers for $x_2$ (\ie, blocks 3-4 marked in gray). This strategy resembles early exiting in unimodal dynamic networks, yet with different motivations. In essence, fusion-level DynMM saves future fusion and modality-wise operations for some multimodal inputs when combining low-level features from each modality (\ie, fusing at early stages) is sufficient for good predictions. On the other hand, for ``hard'' instances, DynMM provides the option of combining multimodal features in each cell with complex fusion operations for maximum representation power. Note that we replace the four individual gating networks $G(\mathbf x)$ in each fusion cell with a global gating network $\hat G(\mathbf x)$ for better integration; $\hat G(\mathbf x)$ takes multimodal features $(x_1, x_2)$ as input and makes decisions on which fusion operation to adopt for the four fusion cells.



This paradigm is especially helpful in tasks where the final prediction is mainly based on a dominant modality, while the other auxiliary modalities provide useful cues to improve the prediction. Fusion-level DynMM provides a flexible way to control \emph{how} and \emph{when} the auxiliary modality comes in to assist the main prediction process. \emph{Progressive fusion} is achieved by our carefully designed fusion cell and dynamic architecture, leading to great computational savings, strong representation power and improved robustness.

Note that modality-level DynMM and fusion-level DynMM are two approaches targeting different granularity levels. In our experiments, we use modality-level DynMM to solve two classification tasks, while the fusion-level DynMM is used for the more challenging semantic segmentation task (\ie, a dense prediction problem).

\subsection{Training Objective}
We notice that for both modality-level and fusion-level DynMM designs, the computation for each expert network $E_i$ (operation $O_i$) is different. Normally, an expert network (an operation) that is computationally heavy has strong representation power. If we directly train the network by minimizing a task-specific loss, the gating network is likely to learn a trivial solution that always chooses the branch with the heavy computation. To achieve efficient inference, we introduce a \emph{resource-aware loss} function into the training objective. Let $C(E_i)$ denote the computation cost (\eg, MAdds) of executing an expert network $E_i$. Similarly, $C(O_{i,j})$ represents the computation cost of the $i$-th fusion operation in the $j$-th cell. Note that the computation cost can be pre-determined before training and is a constant term. The training objectives are shown below:
\begin{gather}
    \mathcal L =  \mathcal L_{task} + \lambda \sum_{i=1}^B g_i C(E_i) \quad \text{(modality-level)}\label{eq.loss1}\\
     \mathcal L =  \mathcal L_{task} + \lambda \sum_{j=1}^F \sum_{i=1}^B g_i^{(j)} C(O_{i,j}) \quad \text{(fusion-level)}\label{eq.loss2}
\end{gather}
where $\mathcal L_{task}$ denotes the \emph{task loss}, \eg, cross entropy between the network prediction and true label for classification. $\mathbf g^{(j)}$ represents the decision vector given by the $j$-th fusion cell. $B$ is the total number of experts (operations) and $F$ is the number of fusion cells. $\lambda$ is a hyperparamter controlling the relative importance of the two loss terms. 

The new objectives (\ref{eq.loss1}) and (\ref{eq.loss2}) account for the computation cost of executing each path and enables DynMM to achieve a desired tradeoff between accuracy and efficiency. We can adjust the value of $\lambda$ based on the deployment constraints. For large $\lambda$, DynMM will prioritize lightweight computations for high computational efficiency. For small $\lambda$, DynMM will explore these computationally heavy paths more often, thus yielding higher accuracy.

\subsection{Optimization}\label{sec.train}
We aim to train DynMM in an end-to-end manner. Since the current gating network provides discrete decisions, the branch selection is not directly differentiable with respect to the gating network. Gumbel-softmax and reparameterization techniques \cite{gumbel} are introduced in the training process. Recall that $\mathbf g$ denotes the desired one-hot $B$-dimensional decision vector produced by a gating network $G(\mathbf x)$, \ie, $\mathbf g = \text{one-hot} (\argmax_i G(\mathbf x)_i)$. We adopt a real-valued soft vector $\mathbf{\tilde g}$ with the following form:
\begin{equation}
    \tilde g_i = \frac{\exp((log G(\mathbf x)_i + b_i)/\tau)}{\sum_{j=1}^B \exp((log G(\mathbf x)_j + b_j)/\tau) }\quad i=1,2,\dots,B
\end{equation}

where $b_1, b_2, \dots, b_B$ are samples independently drawn from Gumbel(0, 1) \cite{gumbel} and $\tau$ denotes the softmax temperature. The distribution of $\mathbf{\tilde g}$ is more uniform with large $\tau$ and resembles a categorical distribution with small $\tau$. $\mathbf{\tilde g}$ serves as a continuous, differentiable approximation of $\mathbf g$. We consider two training techniques: (a) Hard $\mathbf g$ is replaced with soft $\mathbf{\tilde g}$ in Equations (\ref{eq.loss1})-(\ref{eq.loss2}) to enable back-propagation. During training, we anneal $\tau$ so that $\mathbf{\tilde g}$ gradually converges to a desired one-hot vector. (b) Following the straight-through technique \cite{gumbel}, we adopt hard $\mathbf g$ in the forward pass and soft $\mathbf{\tilde g}$ in the backward propagation with the gradient approximation $\nabla \mathbf g \approx \nabla \mathbf{\tilde g}$. In this way, the gating network still outputs a discrete decision during training. Note that we always use hard $\mathbf g$ during inference for computational benefits. Next, we propose a two-stage training of DynMM that jointly optimizes the multimodal network and gating modules.

\textbf{Stage I}: \emph{Pre-training}. We find that following sparse decisions of the gating network in the early stage of training can result in a biased optimization. Branches that are rarely selected have fewer and smaller weight updates; poor performance may result in them getting selected less often (thus never improving). The goal of a pre-training stage is to ensure that every branch of DynMM is fully optimized before the gating modules get involved. For modality-level DynMM, we sufficiently train each expert network at this stage. For fusion-level DynMM, we adopt random decisions (\ie, randomly an operation from the set of candidate operations) for each fusion cell so that each path of the dynamic network is optimized uniformly. 

\textbf{Stage II}: \emph{Fine-tuning}. We incorporate gating networks into our optimization process at this stage. With the reparamterization technique introduced above, we jointly optimize the dynamic network along with gating networks in an end-to-end fashion. 
\section{Experiments} \label{sec.exp}

\subsection{Experimental Setup}
We conduct experiments on three multimodal tasks: (a) movie genre classification on MM-IMDB \cite{imdb}; (b) sentiment analysis on CMU-MOSEI \cite{mosei}; (c) semantic segmentation on NYU Depth V2 \cite{nyu}. To demonstrate the wide applicability of our proposed DynMM, we select the above three tasks that include different modalities (\ie, image and text in task (a), video, audio and text in task (b), RGB and depth images in task (c)). We adopt modality-level DynMM for the first two tasks and fusion-level DynMM for the more challenging semantic segmentation task. Due to space limitations, we present: (1) implementation details; (2) visualization of the gating network decision; (3) an analysis of varying regularization strength $\lambda$; and (4) an ablation study on training strategies of DynMM in the Appendix.

\subsection{Movie Genre Classification}
MM-IMDB is the largest publicly available multimodal dataset for genre prediction on movies. It comprises 25,959 movie titles, metadata and movie posters. We select two movie genres (\ie, drama and comedy) for multi-label
classification from posters (image modality) and text descriptions (text modality). We follow the original data split in \cite{imdb}, and use 15,552 data for training, 2,608 for validation and 7,799 for testing. For preprocessing, we adopt the same method as \cite{imdb,multibench} to extract text and image features.


\begin{table}[t]
\centering
\resizebox{1.0\columnwidth}{!}{
\begin{tabular}{lcccc}
\toprule
Method & Modality & \makecell[c]{Micro\\ F1 (\%)} & \makecell[c]{Macro\\ F1 (\%)} & \makecell[c]{MAdds\\ (M)} \\
\midrule
Image Network & I & 39.99 & 25.26 & 5.0\\
Text Network ($E_1$) & T & 59.16 & 47.21 & 0.7 \\
\hdashline
Late Fusion~\cite{multibench} ($E_2$) & & 59.55 & 50.94 & 10.3 \\ 
LRTF~\cite{lrfusion} & & 59.18 & 49.26 & 10.3 \\ 
MI-Matrix~\cite{jayakumar2020multiplicative}  & & 58.45 & 48.36 & 10.3 \\ 
\midrule
DynMM-a & \multirow{4}{*}{I+T} & \textBF{59.57} & 48.84 & 1.6 \\
DynMM-b & & \textBF{59.59} & 50.42 & 7.8 \\
DynMM-c & & \textBF{59.72} & \textBF{51.20} & 9.8 \\
DynMM-d & & \textBF{60.35} & \textBF{51.60} & 12.1 \\
\bottomrule
\end{tabular}}
\caption{Results on the MM-IMDB Movie Genre Classification. Modalities I and T denote image and text, respectively. The computation cost is measured by multiply-add operations (MAdds) with one image-text pair as the input. M denotes million. Each DynMM variant is obtained using a different value of the regularization hyperparameter $\lambda$ during training.}
\label{tab.imdb}
\end{table}


We adopt two expert networks for this task, namely, a unimodal network $E_1$ that takes textual features as input and another multimodal network $E_2$ that adopts late fusion \cite{multibench} to combine image and text features. We do not consider the use of an image-only network here due to its poor performance on this task. The gating network is a 2-layer MLP with hidden dimension of 128, which takes concatenated image and text features as input and outputs a 2-dimensional vector for expert network selection. We set the temperature of Gumbel-softmax as 1 and adopt straight-through training (\ie, the gating network outputs a one-hot decision vector in the forward propagation). 

Table \ref{tab.imdb} provides the comparison of our proposed modality-level DynMM with static unimodal networks and multimodal networks. We provide results of DynMM under different resource requirements (\ie, use different $\lambda$ in the loss). From Table \ref{tab.imdb}, we can see that DynMM achieves a good balance between computational efficiency and performance. Compared to the static $E_2$ network, DynMM-c improves both MAdds and macro F1 score. DynMM-d provides maximum representation power by using soft gates (which leads to more computation) and achieves best micro and macro F1 scores. On the other hand, DynMM-a involves much less computation, while still maintaining good performance (outperforms $E_1$ by 1.6\% in macro F1). This demonstrates the great flexibility and efficacy of DynMM.




\begin{figure}[!t]
  \centering
  \includegraphics[width=1.0\linewidth]{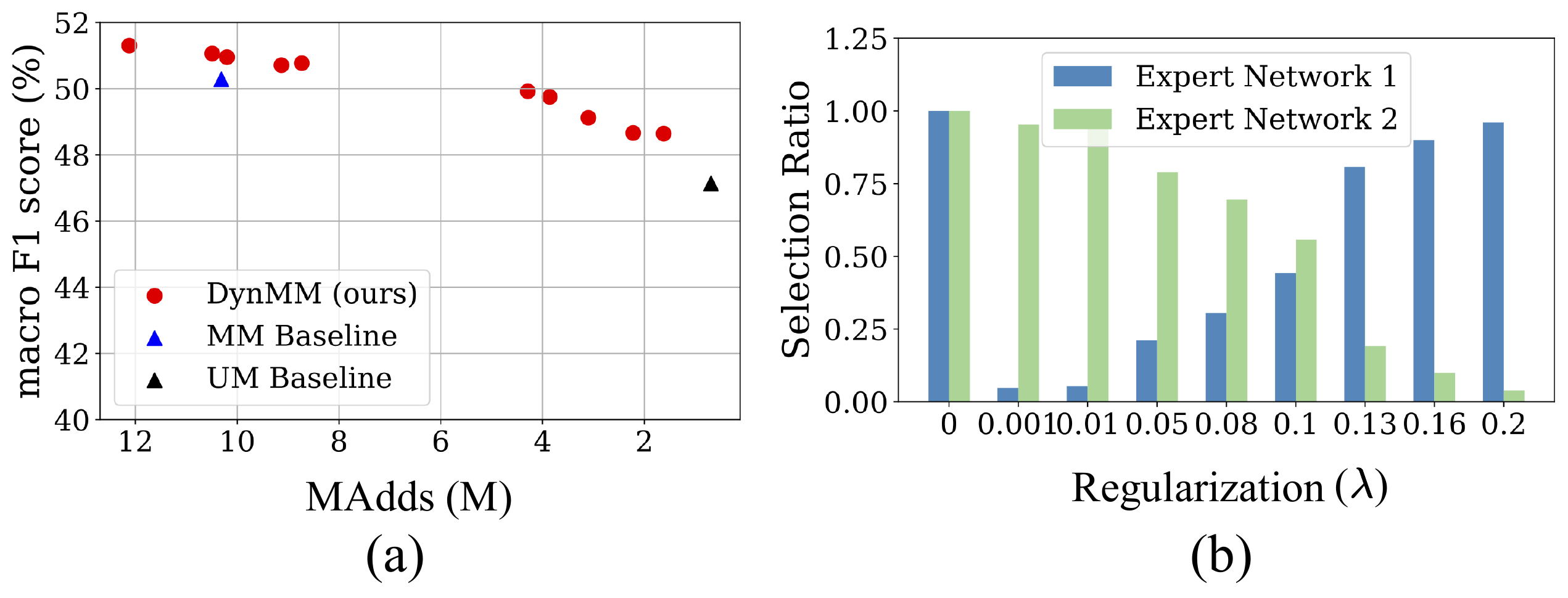}
  \caption{Analysis of DynMM with varying resource regularization strength ($\lambda$) on MM-IMDB. (a): Comparison of DynMM with static unimodal (UM) and multimodal (MM) baselines. (b): Branch selection ratio in DynMM with respect to $\lambda$. DynMM offers a wide range of choices that balance computation and learning behavior well.}
  \label{fig.imdb}
\end{figure}


In addition, we vary $\lambda$ in Equation (\ref{eq.loss1}) to control the importance of resource loss during training. The resulting DynMM models have varying computation costs and performance, as shown in Figure \ref{fig.imdb} (a). On one hand, when compared against a multimodal baseline that is computationally heavy, DynMM maintains good performance with much fewer MAdds. On the other hand, DynMM has better representation power than a unimodal network and thus improves the F1 score. Figure \ref{fig.imdb} (b) shows the selection ratio of each expert network in DynMM with respect to $\lambda$. We observe that as $\lambda$ increases, DynMM focuses more on reducing computation and thus is more likely to select expert network 1 ($E_1$) with a small computation cost. Note that for the $\lambda=0$ case, we adopt soft gates, \ie, every expert network is activated and the output is a weighted combination of predictions given by the two expert networks. Thus, DynMM achieves the best performance at the cost of increased computation. This also demonstrates the flexibility of DynMM, as we can easily adjust $\lambda$ to target high performance or high inference efficiency. 

\subsection{Sentiment Analysis}
CMU Multimodal Opinion Sentiment and Emotion Intensity (CMU-MOSEI) is the largest dataset of sentiment analysis and emotion recognition. It contains 3,228 real-world online videos from more than 1000 speakers and 250 topics.  Each video is split into short segments of 10-20 seconds. Each segment is annotated for a sentiment from -3 (strongly negative) to 3 (strongly positive). The task is to predict the sentiment scores from video, audio and text. Following \cite{multibench}, we use 16,265 data for training, 1,869 data for validation and 4,643 data for testing. The feature extraction steps are the same as \cite{multibench}. 


As text is the best performing modality in this task, we adopt a unimodal network that takes textual features as input to be the expert network $E_1$. The second expert network ($E_2$) of our DynMM is selected as a late fusion network \cite{multibench} that receives inputs from three modalities. The gating network is designed as a lightweight transformer network with hidden dimension equal to 512 and 2 attention heads, followed by a linear layer. The gating network receives concatenated features from three modalities and generates sample-wise decisions on which expert network to activate during inference time. We set temperature of Gumbel-softmax as 1 and adopt straight-through training.

\begin{table}[!tb]
\centering
\resizebox{1.0\columnwidth}{!}{
\begin{tabular}{lcccc}
\toprule
Method & Modality & Acc$^2$ (\%) & MAE & MAdds (M) \\
\midrule
Video Network & V & 69.02 & 0.80 & 123.1 \\
Audio Network & A & 67.68 & 0.82 & 123.3 \\
Text Network ($E_1$) & T & 78.35 & 0.62 & 124.7 \\
\hdashline
Early Fusion~\cite{multibench} &\multirow{2}{*}{V+A+T} & 78.45 & 0.65 & 313.5 \\ 
Late Fusion~\cite{multibench} ($E_2$)  & & 79.54 & \textBF{0.60} & 309.6 \\ 
\midrule
DynMM-a & \multirow{3}{*}{V+A+T} & 79.07 & 0.62 & 165.5 \\
DynMM-b & & \textBF{79.73} & 0.61 & 254.5 \\
DynMM-c & & \textBF{79.75} & \textBF{0.60} & 295.8 \\
\bottomrule
\end{tabular}}
\caption{Results on CMU-MOSEI Sentiment Analysis. Modalities V, A, T represent video, audio and text, respectively. Acc$^2$ denotes binary accuracy (\ie, positive/negative sentiments) and MAE represents mean absolute error. MAdds are measured with a video-audio-text tuple. Each DynMM variant is obtained using a different value of the regularization hyperparameter $\lambda$ during training.}
\label{tab.mosei}
\end{table}

\begin{figure}[!t]
  \centering
  \includegraphics[width=1.0\linewidth]{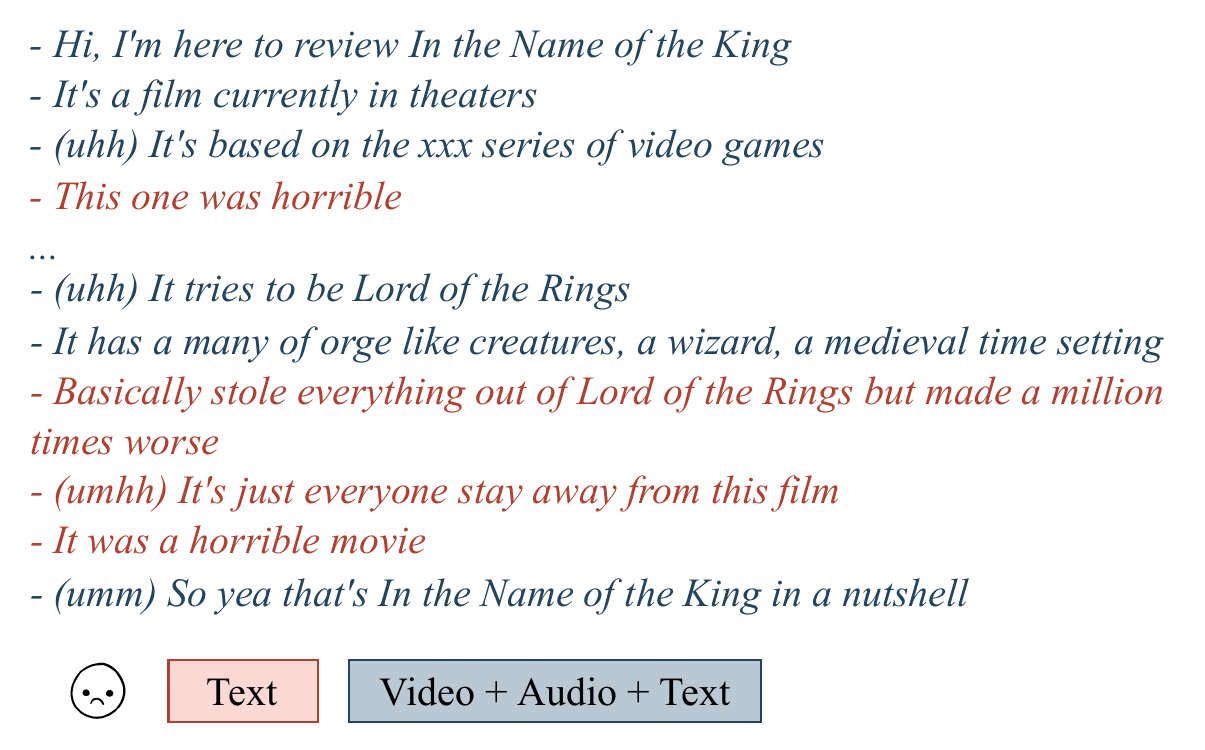}
  \caption{We visualize a few test instances on CMU-MOSEI for a negative sentiment. DynMM identifies sentences marked with red as ``easy'' instances and only uses textual information for prediction. For sentences marked with blue, DynMM takes multimodal inputs (\ie, video+audio+text) for more accurate predictions. }
  \label{fig.viz_mosei}
\end{figure}

Results are summarized in Table \ref{tab.mosei}. We provide three DynMM networks trained with different $\lambda$. Compared with the best performing static network (\ie, Late Fusion), DynMM-a can reduce computations by 46.5\% with a slightly decreased accuracy (\ie, -0.47\%). By allowing more computation, DynMM-b improves both inference efficiency (\ie, reduce MAdds by 17.8\%) and prediction accuracy. Finally, DynMM-c further improves the accuracy by trading off some computation; it achieves best accuracy and smallest mean absolute error with reduced computation cost. These results demonstrate the great advantages of dynamic multimodal fusion. Since multimodal data naturally brings redundancy, we observe that many computations can be reduced without loss in accuracy. 

To have an intuitive sense of our gating network decision on which modality to select, we provide visualization results of several test instances in Figure \ref{fig.viz_mosei}. For simplicity only the text modality is shown here, and the other two modalities (\ie, video and audio) are omitted. The gating network chooses $E_1$ for sentences marked with red and $E_2$ for sentences marked with dark blue. We find that the sentences marked with red often possess strong evidence indicating the sentiments of this sample, \eg, `horrible', `amazingly good'. Therefore, they belong to the ``easy'' samples category that can be correctly predicted using the text modality alone. On the contrary, the sentences marked with dark blue are vague and require additional modalities to help with the prediction. These results indicate that the gating function is well trained and can provide reasonable decisions based on input characteristics.

\subsection{Semantic Segmentation}

NYU Depth V2 is an indoor dataset for semantic segmentation. It contains 1,449 RGB-D images with 40-class labels; 795 images are used for training and 654 images are for testing. The two modalities are RGB and depth images.

\begin{table}[!htb]
\centering
\resizebox{1.0\columnwidth}{!}{
\begin{tabular}{lcccc}
\toprule
Method & \makecell[c]{mIoU\\ (\%)} & \makecell[c]{Depth Enc\\ MAdds (G)} & \makecell[c]{MAdds \\Reduction (\%)} \\
\midrule
ESANet~\cite{esanet} (baseline) & 50.5 & 24.7 & - \\ 
DynMM (Stage I) & 48.5 & 11.7 & 52.6\% \\
DynMM-a (Stage II) & 49.9 & 11.1 & 55.1\% \\
DynMM-b (Stage II) & 51.0 & 19.5 & 21.1\% \\
\bottomrule
\end{tabular}}
\caption{Results on RGB-D semantic segmentation. mIoU denotes mean Intersection-over-Union. MAdds are calculated for input size of $3\times480\times640$. G stands for Giga.}
\label{tab.nyu}
\end{table}

\begin{table}[t]
\centering
\resizebox{1.0\columnwidth}{!}{
\begin{tabular}{lccccc}
\toprule
Method & Modality & Backbone & mIoU (\%) & MAdds (G) \\
\midrule
LW-RefineNet~\cite{lwrefinenet}  & \multirow{2}{*}{RGB} & ResNet-50 & 41.7 & \textBF{38.5}\\ 
LW-RefineNet~\cite{lwrefinenet}  & & ResNet-101 & 43.6  & 61.2\\
\hdashline 
ACNet~\cite{hu2019acnet}  & \multirow{4}{*}{RGB+D} & ResNet-50 & 48.3 & 126.2 \\ 
SA-Gate~\cite{sagate} &  & ResNet-50 & 50.4 & 147.6 \\ 
CEN~\cite{cen}  & & ResNet-101 & \textBF{51.1} & 618.3\\ 
ESANet~\cite{esanet}  &  & ResNet-50 & 50.5 & 56.9 \\
\midrule 
DynMM-a & \multirow{2}{*}{RGB+D} & ResNet-50 & 49.9 & \textBF{43.4} \\ 
DynMM-b & &  ResNet-50 & \textBF{51.0} & 52.2 \\
\bottomrule
\end{tabular}}
\caption{Comparison of our approach with SOTA methods for RGB-D semantic segmentation on NYU Depth V2 test data.}
\label{tab.nyusota}
\end{table}

We adopt fusion-level DynMM for this task and base our dynamic architecture design on a (static) efficient architecture, ESANet \cite{esanet}. As illustrated in Figure \ref{fig.fusion}, we incorporate four fusion cells in the encoder design, where each fusion cell contains two operations. Operation 1 is an identity mapping of RGB features, \ie, $O_1 = x_1$. For the second operation, we use channel attention fusion, where features from both modalities are first reweighted with a Squeeze and Excitation module \cite{hu2018squeeze} and then added element-wisely. Two ResNet-50 \cite{resnet} are used as feature extraction models for RGB and depth modality. The decoder design is identical to \cite{esanet}. The gating network comprises a pipeline of 2 convolution blocks with kernel size 5$\times$5 and stride size 2, a global average pooling and a linear layer. RGB and depth features after the first convolutional layer are concantenated together and passed to the convolutional gate. The gating network outputs a 4-dimensional vector per sample that determines which operation to select for each fusion cell. We experiment with two training strategies: (1) DynMM-a in Table \ref{tab.nyu} is trained with straight-through technique with Gumbel-softmax temperature $\tau=1$; (2) We obtain DynMM-b in Table \ref{tab.nyu} by exponentially decaying $\tau$ from 1 to 0.0001 during 500 epochs.

Table \ref{tab.nyu} provides the detailed results of fusion-level DynMM. We report performance of DynMM after first-stage training in the second row; its great performance validates the design of our random gating function in the pre-training stage. This also lends support to our claim that there exists a lot of redundancy in multimodal networks. Utilizing the finding that depth modality plays an auxiliary role in this task, fusion-level DynMM effectively reduces computations of the depth encoder. DynMM-a reduces MAdds by 55.1\% with only -0.4\% mIoU drop. Furthermore, DynMM-b achieves a mIoU improvement of 0.7\% and 21.1\% reduction in MAdds at the same time, thus demonstrating the superiority of DynMM over static fusion.

Table \ref{tab.nyusota} presents a comparison of the resulting DynMM-a and DynMM-b with SOTA semantic segmentation methods. For baseline methods, we list mIoU reported in their original papers and report MAdds. These results clearly show that our proposed method achieves the best balance between performance and efficiency. The computation cost of DynMM is similar to a unimodal lightweight RefineNet, yet its performance can match methods that use ResNet-101 as the backbone and involve significantly larger MAdds. 

\begin{figure}[!t]
  \centering
  \includegraphics[width=1.0\linewidth]{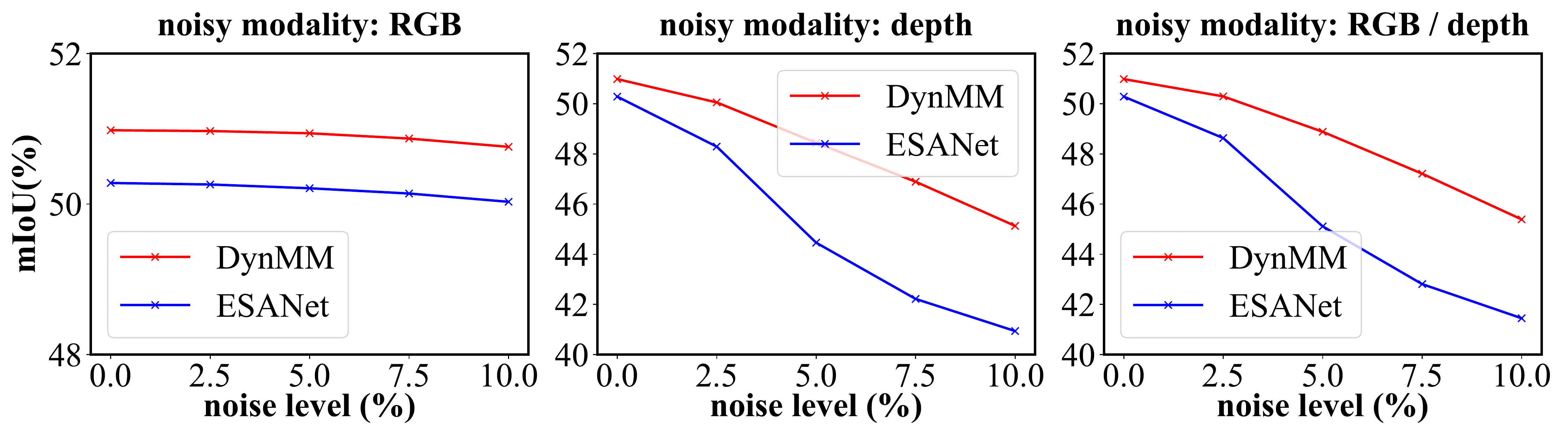}
  \caption{DynMM vs. ESANet on NYU Depth V2 with different degrees of Gaussian noise injected into RGB / depth images. }
  \label{fig.noise}
\end{figure}
\begin{figure}[!t]
  \centering
  \includegraphics[width=1.0\linewidth]{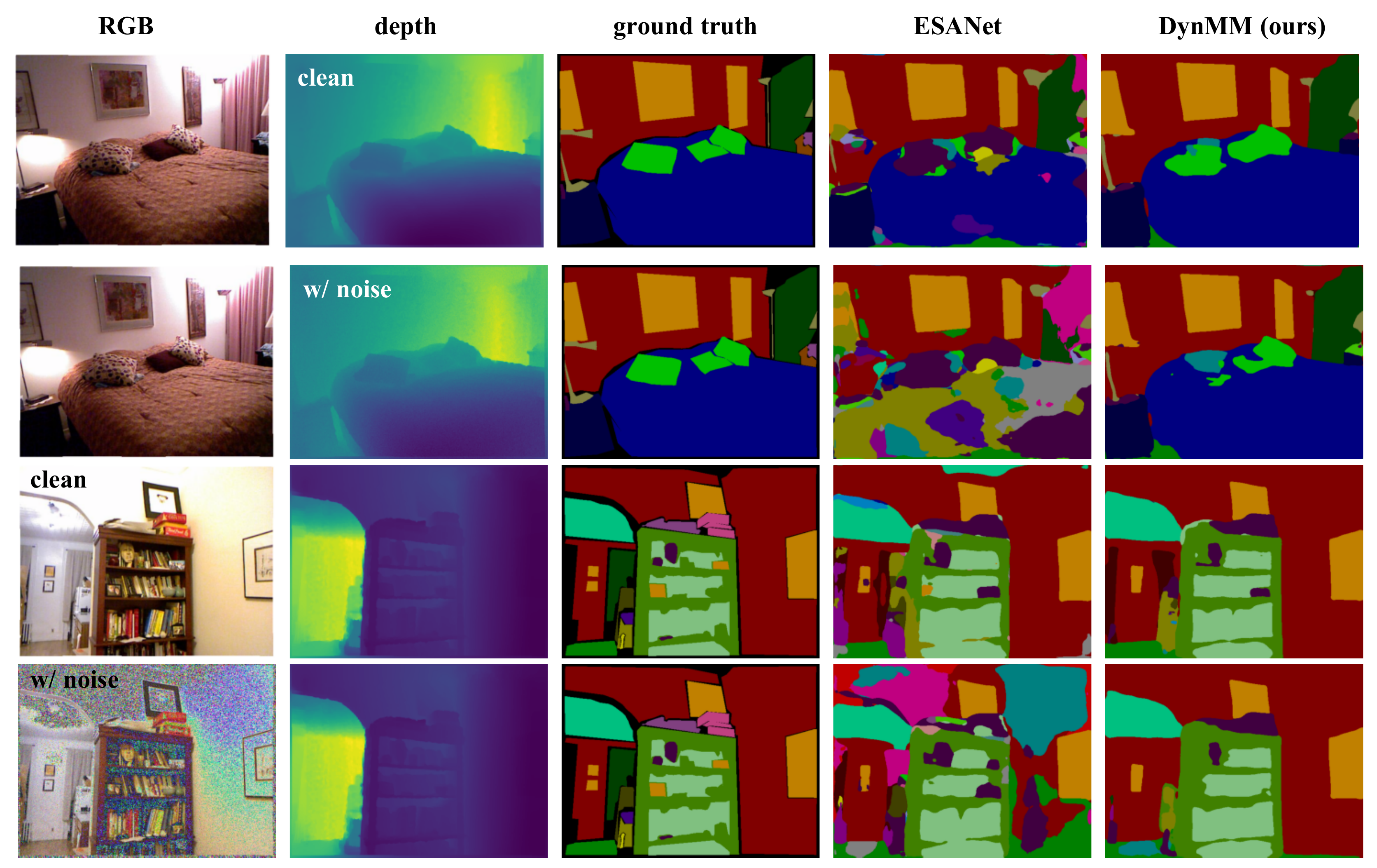}
  \caption{Qualitative segmentation results on NYU Depth V2. DynMM is more robust to noisy multimodal data compared with the static ESANet.}
  \label{fig.seg}
\end{figure}

Finally, we conduct experiments to demonstrate the improved robustness of DynMM compared to ESANet. We consider three settings by injecting random Gaussian noise with probability 1/3 to (1) RGB modality; (2) depth modality and (3) both modalities. We experiment with different degrees of random Gaussian noise and plot the performance degradation of two approaches in Figure \ref{fig.noise}. From the figure, we observe that the performance gap between DynMM and ESANet becomes larger when the noise level of depth images increases; This demonstrates another advantage of DynMM in reducing data noise and improving robustness. Figure \ref{fig.seg} shows some qualitative segmentation results. While ESANet generates reasonable predictions in the normal setting (\ie, first and third row), its performance becomes significantly worse when multimodal data is perturbed by noise (\ie, the second and fourth row). On the contrary, our DynMM is robust to noise and provides a good prediction for both scenarios. These results suggest the potential of a dynamic neural network architecture for improving robustness of multimodal fusion.

\section{Conclusion} \label{sec.conclusion}
Multimodal data enable models to learn from an enriched representation space, but it also bring significant redundancy. Motivated by this observation, we have proposed dynamic multimodal fusion (DynMM), a new approach that adaptively fuses inputs during inference. 
Experimental results on three very different multimodal tasks demonstrate the efficacy of DynMM. More importantly, our work demonstrates the potential of dynamic multimodal fusion and opens up a new research direction. Considering the benefits of a dynamic architecture (\ie, reduced computation, improved performance and robustness)
, we believe that developing dynamic networks tailored for multimodal fusion is a topic worthy of further investigations. 

DynMM has limitations that we plan to address through three areas of improvement in our future work. These include designing better dynamic architectures that can account for multimodal redundancy, extending DynMM to sequential decision-making tasks, such as long video prediction and exploring the performance of DynMM on different multimodal tasks and modalities.


{\small
\bibliographystyle{ieee_fullname}
\bibliography{egbib}
}
\newpage
\appendix
In this appendix, we present: (1) implementation details; (2) visualization results of decisions given by the gating network (on NYU Depth V2); (3) an analysis of varying regularization strength $\lambda$ (on CMU-MOSEI); and (4) an ablation study on proposed training strategies (on NYU Depth V2).

\section{Implementation Details}
\textbf{MM-IMDB}. $E_1$ is a unimodal text network with 2-layer MLPs (hidden dimension=512) as the text encoder and the decoder. $E_2$ is a multimodal late fusion network, where we use the text and image encoders to extract features, concatenate the unimodal features and then pass the concantenated features to a MLP decoder (hidden dimension=1024). The text encoder is the same as in $E_1$ and the image encoder is a 2-layer MLP (hidden dimension=1024).  We use AdamW optimizer with lr=1e-4 and weight decay=1e-2.

\textbf{CMU-MOSEI}. $E_1$ is a text network consisting of a 5-layer transformer encoder (hidden dimension=120; 5 attention heads) and a 2-layer MLP decoder (hidden dimension=64). $E_2$ is a multimodal late fusion network with video, audio, and text encoders being 5-layer transformers and a 2-layer MLP decoder (hidden dimension=128). We use AdamW optimizer with lr=1e-4 and weight decay=1e-4.

\textbf{NYU Depth V2}. The image and depth encoder is a ResNet-50 and the decoder is the same as in ESANet~\cite{esanet}. We use SGD optimizer with weight decay=1e-4 and momentum=0.9, also OneCycleLR with max\_lr=1e-2. 

The gating networks are designed to match the $E_1$ and $E_2$ model architectures. Therefore, we use a MLP gate for MM-IMDB, a transformer gate for CMU-MOSEI and a convolution gate for NYU Depth V2. 

$C(E_i)$ in Equations (\ref{eq.loss1})-(\ref{eq.loss2}) is set as the MACs required to do one forward pass with $E_i$. Take MM-IMDB for example: the MACs for executing $E_1$ and $E_2$ are 1.25M and 10.87M, respectively. The resource loss of one data sample is $\lambda$ if the gating network selects $E_1$ and $\lambda \times \frac{10.87}{1.25}$ if $E_2$ is selected. The DynMM variants reported in Table \ref{tab.imdb}-\ref{tab.mosei} are obtained using different values of the regularization parameter $\lambda$. 

\section{Visualization Results}

In our proposed DynMM, the gating network is crucial as it provides data-dependent decisions on which expert network to adopt. For modality-level DynMM, we have provided visualization of  the gating network decisions for some test instances on CMU-MOSEI in Figure \ref{fig.viz_mosei} in the main paper. Similarly, for fusion-level DynMM, we visualize several test instances on NYU Depth V2 and the resulting architecture in Figure \ref{fig.viz_nyu} in the Appendix. 

From Figure \ref{fig.viz_nyu}, we can see that DynMM adaptively executes the forward path for multimodal inputs. The depth features are combined with the RGB features to different degrees, determined by the gating network in DynMM. This provides a flexible way to control multimodal fusion in a sample-wise manner. For the RGB-D images in the upper figure, DynMM performs one-time fusion for multimodal features after the first block and saves computations of depth blocks 2-4. For the more challenging test samples in the lower figure, DynMM decides to fuse features in every layer to better incorporate multimodal information. Due to the dynamic architecture, DynMM achieves a good balance between efficiency and performance.

\section{Analysis of Regularization Strength}

\begin{figure}[!t]
  \centering
  \includegraphics[width=1.0\linewidth]{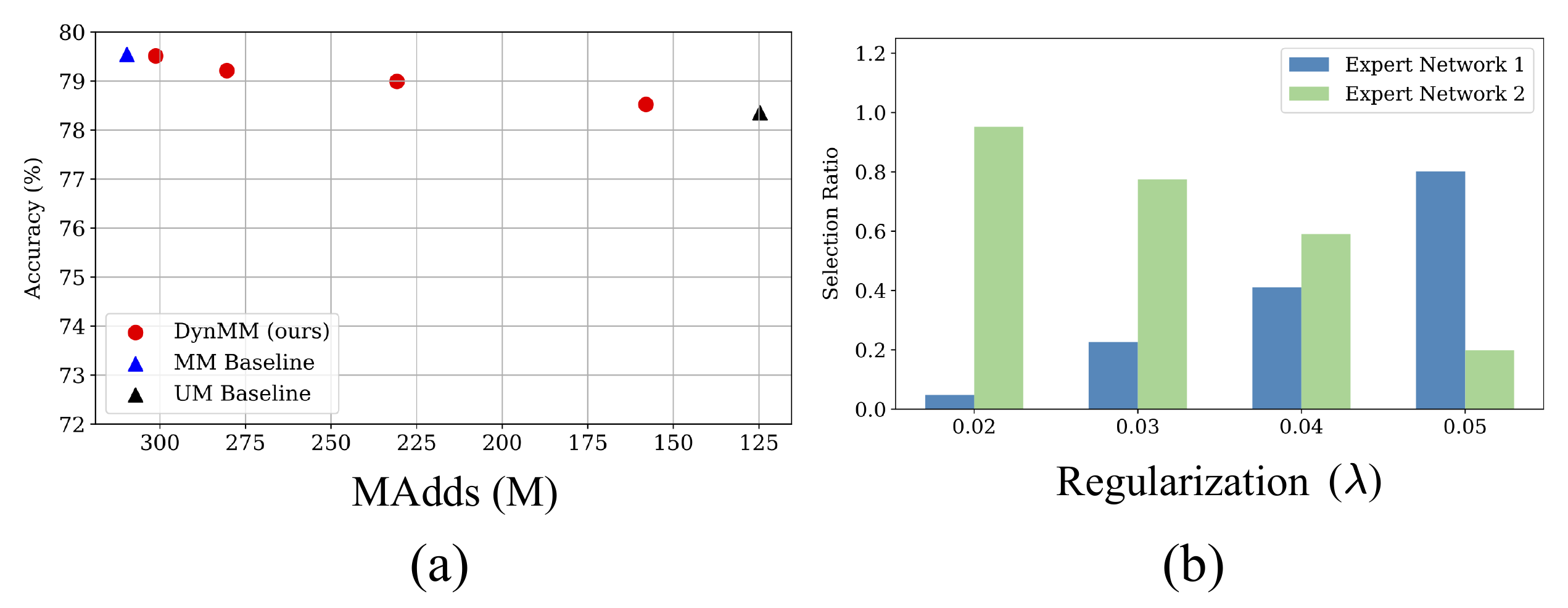}
  \caption{Analysis of DynMM with varying resource regularization strength ($\lambda$) on CMU-MOSEI. (a): comparison of DynMM with static unimodal (UM) and multimodal (MM) baselines. (b): branch selection ratio in DynMM with respect to $\lambda$.}
  \label{fig.mosei}
\end{figure}

Recall that we propose a resource-aware loss function in Equation (\ref{eq.loss1}) and (\ref{eq.loss2}) in the main paper, where $\lambda$ is a hyperparameter controlling the relative importance of task loss and computation cost loss. Similar to Figure \ref{fig.imdb} in the main paper (\ie, an analysis of $\lambda$ on MM-IMDB), we vary $\lambda$ when training DynMM on CMU-MOSEI sentiment analysis and report its computation cost and performance corresponding to each $\lambda$ value. The results are provided in Figure \ref{fig.mosei} in this Appendix. From Figure \ref{fig.mosei} (a), we can see that DynMM achieves a good balance between inference efficiency and accuracy. Moreover, DynMM offers a wide range of choices that can be controlled by $\lambda$, thus showing great flexibility. Figure \ref{fig.mosei} (b) shows the branch selection ratio of DynMM for different $\lambda$. When $\lambda$ is small, DynMM focuses more on performance and chooses expert network 2 most of the time. As $\lambda$ increases, more test samples are routed to the expert network 1 that requires fewer computations.

\begin{figure}[!t]
    \centering
    \includegraphics[width=1.0\linewidth]{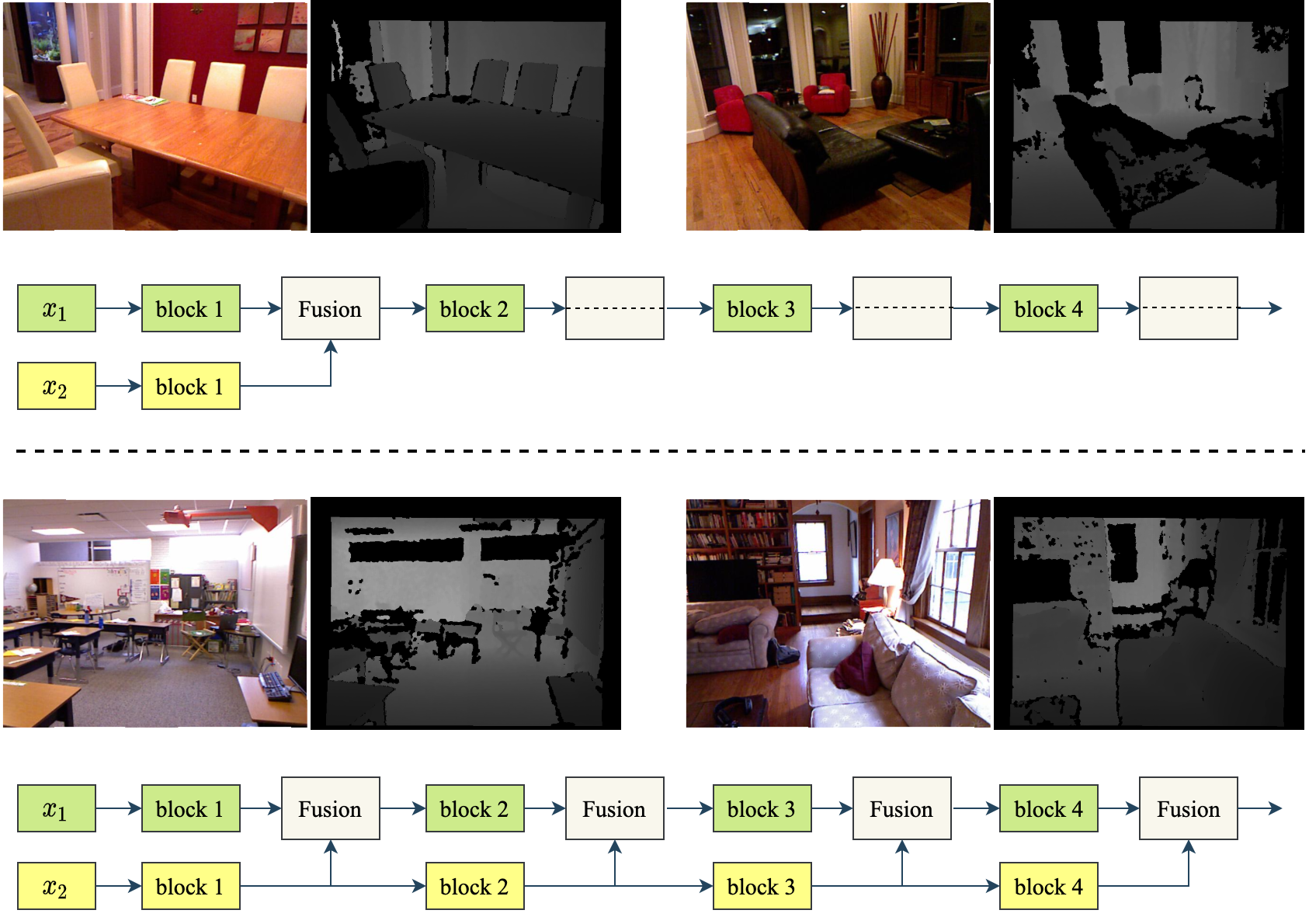}
    \caption{We visualize a few test instances on the NYU Depth V2 data. $x_1$ and $x_2$ denote RGB and depth images, respectively. The corresponding network architecture based on the gating network decision is shown. The upper figure shows examples when the gating network chooses an early fusion architecture. DynMM skips computations of the depth extraction layers, thus achieving inference savings. The lower figure shows examples when the gating network decides to fuse representations at every middle layer. }
    \label{fig.viz_nyu}
\end{figure}

\section{Ablation Study}

\begin{table}[!b]
\begin{center}
\scalebox{0.9}{
\begin{tabular}{cccc}
\toprule
Method & \makecell[c]{Two-stage\\ Training} & \makecell[c]{Joint\\ Optimization} & \makecell[c]{mIoU (\%)} \\
\midrule
Baseline & & & 50.3  \\
\hdashline
\multirow{3}{*}{DynMM} & & $\checkmark$ & 49.2  \\ 
& $\checkmark$ & & 50.2  \\
 & $\checkmark$ & $\checkmark$ & 51.0  \\
\bottomrule
\end{tabular}
}
\caption{Ablation study on RGB-D semantic segmentation. Baseline refers to a static model (ESANet).}
\label{tab.nyuablation}
\end{center}
\end{table}

To verify the efficacy of our proposed training strategies, we present an ablation study of RGB-D semantic segmentation on the NYU Depth V2 data. We train DynMM under three settings: (1) We omit the pre-training stage and train DynMM in one stage. (2) In the second stage of training, we freeze the weights of the multimodal architecture and only fine-tune the gating network. (3) We adopt our proposed two-stage training with joint optimization of the multimodal network and gating network. The other training parameters (\eg, learning rate, resource regularization strength $\lambda$) are identical. The results are shown in Table \ref{tab.nyuablation} below.

Table \ref{tab.nyuablation} demonstrates the advantages of our proposed training strategies. We observe that DynMM with one-stage training does not have a dynamic architecture, \ie, all test samples are routed to one particular forward path. Without a pre-training stage, every forward path is not equally optimized. Biased optimization further leads to suboptimal performance (\ie, an mIoU of 49.2\%). Apart from two-stage training, joint optimization also plays an important role. We observe a +0.8\% mIoU improvement with end-to-end training. The possible reason is that (static) feature extraction layers also improve in the joint optimization process; they provide more informative features as input to the gating network to a better gating network decision. Therefore, the joint optimization achieves the overall best performance.
\end{document}